\documentclass[runningheads]{llncs}

\usepackage{eccv}

\usepackage{eccvabbrv}

\usepackage{graphicx}
\usepackage{booktabs}
\usepackage{wrapfig}

\newcommand{\figref}[1]{Fig.~\ref{#1}}
\newcommand{\tabref}[1]{Table~\ref{#1}}

\usepackage{color,colortbl}

\definecolor{Gray1}{gray}{0.95}
\definecolor{Gray2}{gray}{0.99}
\newcommand{\CC}[1]{\cellcolor{gray!#1}}
\newcommand\CCG[1][]{\CC{10}}
\usepackage{enumitem, lipsum, calc}
\usepackage[dvipsnames]{xcolor}
\usepackage[accsupp]{axessibility}  %

\usepackage[pagebackref,breaklinks,colorlinks,citecolor=eccvblue]{hyperref}

\definecolor{grn}{rgb}{0, 0.6, 0}
\usepackage{orcidlink}

\begin{document}

\title{Weak-to-Strong Compositional Learning from Generative Models for Language-based Object Detection} 

\titlerunning{Weak-to-Strong Compositional Learning from Generative Models}

\author{Kwanyong Park\inst{1} \and
Kuniaki Saito\inst{2} \and
Donghyun Kim\inst{3}\thanks{Corresponding author}}

\authorrunning{K. Park et al.}

\institute{ETRI \and OMRON SINIC X Corporation \and Korea University}

\maketitle

\begin{abstract}
  Vision-language (VL) models often exhibit a limited understanding of complex expressions of visual objects (\eg, attributes, shapes, and their relations), given complex and diverse language queries. Traditional approaches attempt to improve VL models using hard negative synthetic text, but their effectiveness is limited. In this paper, we harness the exceptional compositional understanding capabilities of generative foundational models. We introduce a novel method for structured synthetic data generation aimed at enhancing the compositional understanding of VL models in language-based object detection. Our framework generates densely paired positive and negative triplets (image, text descriptions, and bounding boxes) in both image and text domains. By leveraging these synthetic triplets, we transform `weaker' VL models into `stronger' models in terms of compositional understanding, a process we call ``Weak-to-Strong Compositional Learning'' (WSCL). To achieve this, we propose a new compositional contrastive learning formulation that discovers semantics and structures in complex descriptions from synthetic triplets. As a result, VL models trained with our synthetic data generation exhibit a significant performance boost in the Omnilabel benchmark by up to +$5$AP and the D$^{3}$ benchmark by $+6.9$AP upon existing baselines.
\end{abstract}

\section{Introduction}
\label{sec:intro}

Recently, vision-language (VL) models have demonstrated significant advancements in visual recognition by learning from large-scale weakly supervised image-text pair datasets~\cite{clip,align}. While traditional recognition models~\cite{fasterrcnn,coco,objects365,lvis} are restricted to classifying or detecting pre-defined classes, image-text paired data allow models to easily generalize to new concepts and domains with language queries.
For example, GLIP~\cite{glip} can perform phrase grounding or detect multiple objects in language queries by learning to align words and regions in each modality.

\begin{figure*}[t]
    \centering 
    \includegraphics[width=1\textwidth]{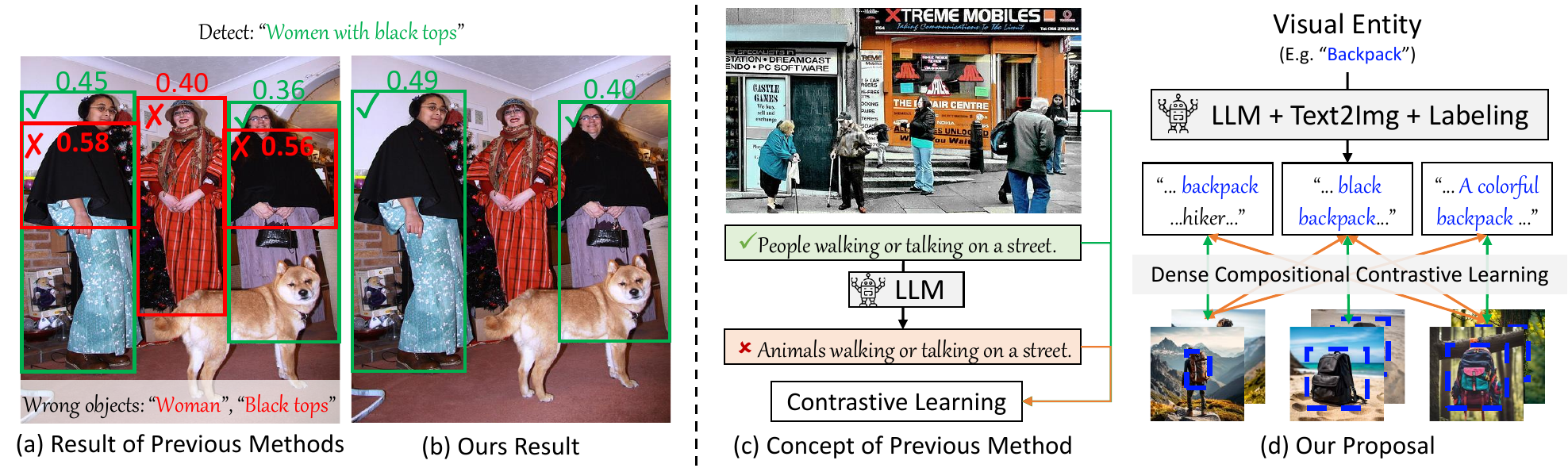}
    \captionsetup{font=footnotesize}
    \caption{
    (a-b) While previous models lack a compositional understanding of the given language query and localize wrong objects, resulting in higher scores for the wrong objects (\ie, the middle woman and black tops) than the actual object, our method successfully localizes only correct objects corresponding to the query. (c) Previous VL methods apply (\eg, \cite{svlc,desco}) augmentation exclusively in the text domain. (d) The proposed method produces comprehensive synthetic triplets comprising <image, object description, bounding box>, incorporating compositional contrastive learning to improve the model’s understanding of composition.}
    \label{fig:motivation}
\end{figure*}

Despite advancements, VL models~\cite{clip,glip} continue to face challenges in understanding complex language queries and structured vision-language concepts, such as detailed object attributes, shapes, textures, and their relationships~\cite{ winoground,aro,svlc}.
A recent study~\cite{aro} indicates that VL models often function like bags-of-words, lacking compositional understanding. This results in a significant performance drop in image-text retrieval tasks involving complex scenes and detailed captions with rich compositional structures.
In the context of object detection, novel benchmarks like OmniLabel~\cite{omnilabel} and D$^{3}$~\cite{d3} have been introduced to assess the ability to interpret a broad range of complex object descriptions and accurately detect target objects (See \figref{fig:motivation}-(a)). 
In such scenarios, VL models frequently overlook the complex and free-form textual descriptions provided, leading to incorrect detection results.
To address this issue, previous work~\cite{desco} has explored augmenting the text domain~\cite{winoground,aro,svlc} by generating synthetic negative texts through noun swapping or creating new image captions (See \figref{fig:motivation}-(c)). 
However, we observe that merely enriching the text domain is insufficient for models to learn dense relations between images and text.

To this end, we propose an innovative framework to distill the unprecedented compositional understanding of recent generative foundational models, such as large language models~\cite{GPT3,GPT4,touvron2023llama} and text-to-image diffusion models~\cite{rombach2022high,sdxl,sdxl_turbo,lee2023koala,pixart}, into VL models. Within our framework, a series of generative models automatically generates synthetic data, from which a language-based object detector learns and inherits compositionality. Through this process, `weaker' VL models evolve into `stronger' models in terms of compositional understanding; we term this process ``Weak-to-Strong Compositional Learning'' (WSCL). \figref{fig:motivation}-(d) illustrates the proposed framework.

To be specific, our framework consists of two steps: (1) Generating diverse and dense triplets. Instead of solely relying on difficult-to-obtainable real-world data, we propose to generate dense triplets (\ie, <image, object description, bounding box>) with the generative models (Sec.~\ref{method:synthetic_generation}). We first use a large language model to collect diverse and dense variations of visual entities (e.g., attributes, relations) in the text domain, then translate these descriptions to the image domain with the text-to-image diffusion models. As a last piece, we localize depicted visual entities as a bounding box. In this step, we decompose the hard grounding problem into multiple easy detection problems, and this simple yet effective change enables us to obtain an accurate bounding box. 
Note that our generation framework is scalable due to its automatic data construction process.
(2) Effective learning from densely generated triplets (Sec.~\ref{method:training_w_synthetic}).
For an image of a specific visual entity, we first contrast the dense variation of descriptions and the detector is trained to detect the object only for the corresponding descriptions. This forces the detector to be aware of the given descriptions.
Besides, we use structural information in the textural description to identify the subject entity and use it to suppress the predictions for the non-subject entities in the descriptions. 
Both contrastive learning method largely improves compositional understanding, resulting in significant performance gain in description-based object detection.
We call the two synergetic contrastive learning methods as compositional contrastive learning.

We utilize our method to enhance two advanced language-based object detection models, namely GLIP~\cite{glip} and FIBER~\cite{fiber}. 
On the challenging Omnilabel benchmark~\cite{omnilabel}, our proposal achieves a notable improvement of +5.0 and +4.8AP upon GLIP-T and FIBER-B.
This suggests that our method effectively enhances its compositional understanding of visual objects and descriptions.
Specifically, for long queries, the performance of the GLIP-T model is doubled from 8.2 to 16.4AP.
Besides, our proposal is proven to be complementary to the previous text augmentation-based method, DesCo~\cite{desco}, and achieves the new state-of-the-art. \textbf{Our contribution} can be summarized as follows: 
\begin{enumerate}[leftmargin=*, topsep=1mm]
\item To our knowledge, this is the first work to generate diverse and dense synthetic triplets for language-based object detection, which are hard to obtain without expensive human annotations. 
\item We present a novel compositional contrastive learning approach that efficiently learns to comprehend intricate compositions in images and text, and aligns image regions with the correct textual descriptions.
\item Our method is model-agnostic and can be applied to diverse prior language-based object detectors. We show that our method significantly improves the performance of the prior detectors on the two challenging benchmarks, OmniLabel and D$^{3}$, across diverse model architectures. 
\end{enumerate}

\section{Related Work}

\noindent\textbf{Vision-language Models.} Vision-language (VL) models (\eg, CLIP~\cite{clip}, ALIGN\cite{align}, GLIP~\cite{glip}) shows remarkable progress in diverse visual recognition tasks. CLIP and ALIGN are pre-trained on large-scale weakly supervised image-text pairs collected from the web with image-level contrastive learning objectives. In order to gain a fine-grained understanding of images, several methods such as GLIP~\cite{glip} propose region-level contrastive alignment between image regions and words in the text. GLIP additionally leverages detection and phrase grounding benchmarks and enables context-free object detection with language queries. However, as studies in \cite{winoground,aro,svlc,zhao2022vl,ma2023crepe,hsieh2024sugarcrepe}, VL models exhibit a limited compositional understanding of complex scenes and rich text descriptions for object attributes, texture, and their relations. In order to address these, hard negative and positive augmentation techniques on the language domain have been proposed in \cite{svlc,aro,dac} and improve its ability of compositional understandings.  On the other hand, we propose to generate synthetic triplets including synthetic data in both image and text domains, and automatically generate bounding boxes for language-based object detection.

\noindent\textbf{Object Detection.} Traditional detection models are trained to detect objects for a pre-defined set of categories~\cite{fasterrcnn,detr,woo2022bridging,redmon2016you}. As a result, traditional models find it challenging to adapt to new tasks and domains, unable to differentiate between objects that vary in attributes such as texture, shape, and other characteristics. Recently, language-based object detection with vision-language models has demonstrated significant potential to enhance their adaptability by utilizing language queries. CLIP~\cite{clip} opens a new research direction in open-vocabulary object detection~\cite{vild,mdetr,kim2023region,zhou2022detecting} demonstrating strong performances on unseen categories by leveraging text encoders like BERT~\cite{bert}. MDETR~\cite{mdetr} detects objects conditioned on complex language queries containing object attributes and relations. However, MDETR struggles to perform effectively on the Omnilabel benchmark~\cite{omnilabel}, which presents queries with more intricate and challenging negative descriptions in free-form text. DesCo~\cite{desco} employs large language models to generate synthetic rich language descriptions to improve the compositional understanding of language queries. Conversely, our research focuses on enhancing language-based object detection by utilizing synthetic triplets that incorporate pseudo bounding boxes for every object description. This is a significant challenge as existing detectors lack compositional understanding. To address this, we transform the complex task into several simpler detection tasks, thereby achieving precise bounding boxes for each description.

\noindent\textbf{Learning from Synthetic Data.} Deep learning models require massive labeled data to obtain strong performances. However, it is expensive to collect such labeled data. On the other hand, synthetic data can be obtained easily to train a model. Learning from synthetic data has been an active research topic for many years in diverse computer vision applications such as image classification~\cite{task2sim,gan2020threedworld,mikami2021scaling,xu2021cdtrans,hur2023learning}, object detection~\cite{peng2015learning,prakash2019structured,hsu2020every,lin2023explore}, and image segmentation~\cite{gta5,tsai2018learning,park2020discover,wang2020differential,shin2021unsupervised,park2023mask}. These models utilize graphics engines to generate images, which causes a domain gap from real data. Recently, several works utilize text-to-image diffusion models~\cite{rombach2022high,sdxl} to generate synthetic images for visual recognition~\cite {azizi2023synthetic,he2022synthetic,tian2024stablerep,fan2024scaling,tian2024learning,nguyen2024dataset,wu2023datasetdm}. However, in our experiments, naively adding synthetic data as a set of training data does not necessarily improve the compositional reasoning ability of VL models. Therefore, we introduce a new compositional contrastive learning that effectively utilizes synthetic image-text paired data for our task.

\section{Background: Language-based Object Detection}
Language-based object detection takes free-form language queries and an image as inputs to identify and predict bounding boxes, aligning these boxes with the corresponding language queries. The task encompasses an open-set and multi-label framework, where queries may include descriptions of objects with unseen and intricate compositions~\cite{omnilabel,d3}. Furthermore, the descriptions may correspond to zero, one, or several instances within the image, diverging from typical object detection~\cite{coco,objects365}. Such characteristics require a VL model to understand complex compositions in visual scenes and textual descriptions. 

 Several VL models (\eg, GLIP~\cite{glip} and FIBER~\cite{fiber}) are utilized to solve this task. We review GLIP~\cite{glip}, and its approach to addressing this task. GLIP redefines detection as a grounding task by matching each region or box $B$  in an image $I$ with phrases in a text query (prompt) $Q$ with a target alignment label $T$. The key is to transform existing data into a grounding format. For detection data, the query $Q$ contains a list of pre-defined object classes such as "Person. Bicycle, ..., glasses". For image-text paired data (\eg, CC12M, SBU~\cite{cc12m,sbu}), the query is a text caption containing entities in the image. Since $T$ is not available for this data, GLIP generates pseudo-grounding labels for the alignment between entities in the caption and regions in the image. Then a model is trained to align each word in the query $Q$ with each region $B$ in (pseudo) $T$ as follows:
 \begin{equation}
     O, P = G(I, Q), \quad S_{\text{ground}} = OP^{\top}, \quad \mathcal{L} = \mathcal{L}(S_{\text{ground}}, T) + \mathcal{L}_{loc}
 \end{equation}
where $G$ is the GLIP model, $O \in \mathbb{R}^{N\times d}$ are the regions features of $I$, $P \in \mathbb{R}^{M \times d}$ is the contextual word tokens features of $Q$,  and $S_{\text{ground}}\in \mathbb{R}^{N \times M}$ is the alignment score. GLIP is trained to minimize the region-word matching and localization loss as in the standard object detection. GLIP struggles to identify the correct region in response to a complex query and fails to generate precise labels.

\section{Method}
We aim to improve the compositional understanding capabilities of a language-based object detector. Instead of relying on difficult-to-obtain triplets (image, object descriptions, and bounding boxes), we harness the capabilities of foundational models by generating these triplets as training samples. Our approach involves two main steps: (1) dense synthetic triplet generation (Sec.~\ref{method:synthetic_generation}) and (2) compositional contrastive learning with dense synthetic triplets (Sec.~\ref{method:training_w_synthetic}). In the first step, we introduce our method to generate diverse and semantically rich training triplets (\ie, objects, object descriptions, and bounding boxes) in both image and text domains. Subsequently, we introduce compositional contrastive learning to effectively improve compositional understanding of visual objects and align with its complex object descriptions from our generated triplets for our language-based object detector.

\begin{figure*}[t]
    \centering 
    \includegraphics[width=1\textwidth]{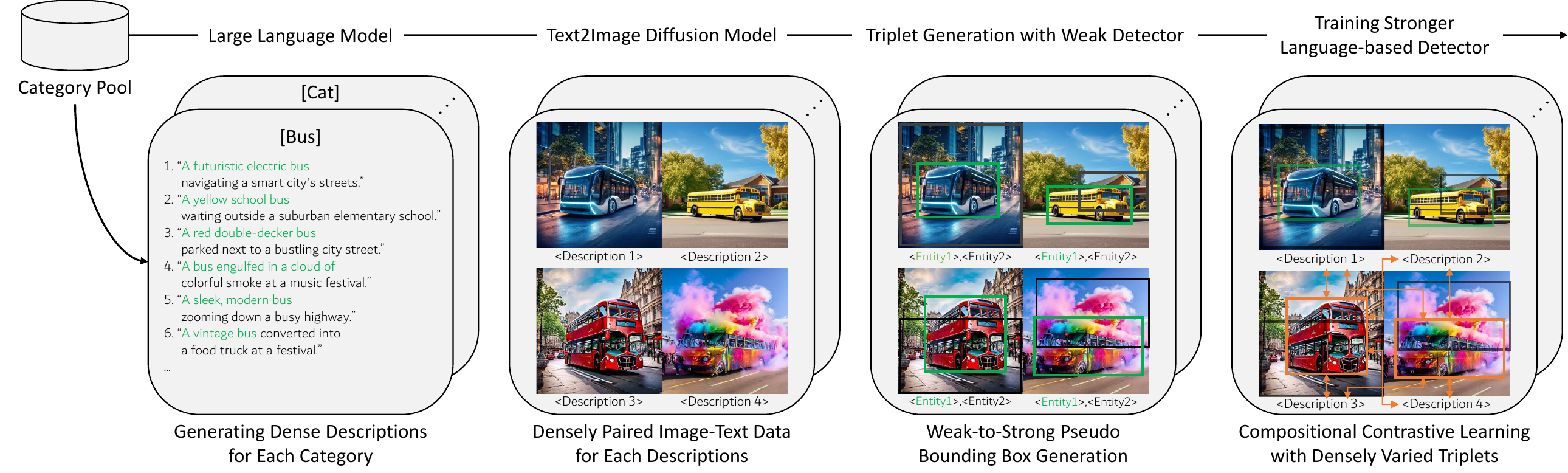}
    \captionsetup{font=footnotesize}
    \vspace{-5mm}
    \caption{
    Overview of our method. Our method consists of generating dense synthetic image-text paired triplets with generative models and creating bounding boxes. Finally, we introduce compositional conservative learning with our generated triplets which enhances the model's compositional ability in language-based object detection.}
    \vspace{-4mm}
    \label{fig:framework}
\end{figure*}

\subsection{Synthetic Triplet Generation in Image and Text Domains}
\label{method:synthetic_generation}

A traditional training data collection process for grounding data~\cite{coco,flickr30k} is to \textit{collect images}, and manually \textit{annotate object bounding boxes with their text descriptions}. However, it would be prohibitively expensive to manually collect images that cover the full diversity of objects along with all their possible attributes, actions, and interactions with their environment. Additionally, localizing these objects and providing free-form textual descriptions of them is even more challenging. And, descriptions provided by human annotators are often brief and lack detail, which can hinder the effective learning of visual-language alignment. Furthermore, this process does not guarantee to obtain hard negatives (\ie, dense triplets) which is crucial to improve the compositionality of VL models~\cite{svlc,desco}. In order to obtain diverse and dense triplets, we adopt a reversed approach which first \textit{generates text descriptions} and then \textit{collects corresponding images}: we begin by synthesizing diverse and plausible object descriptions, then proceed to generate corresponding images inspired by the recent breakthroughs in foundation models~\cite{bert,GPT3,GPT4,rombach2022high,sdxl,sdxl_turbo,lee2023koala,pixart}, and finally, automatically localize the objects within these images. The overview of the proposed training data generation framework is depicted in \figref{fig:framework}. As a result, our method allows the automatic generation of dense triplets without requiring human annotation of text descriptions and bounding boxes.

\begin{figure*}[t]
    \centering 
    \includegraphics[width=1\textwidth]{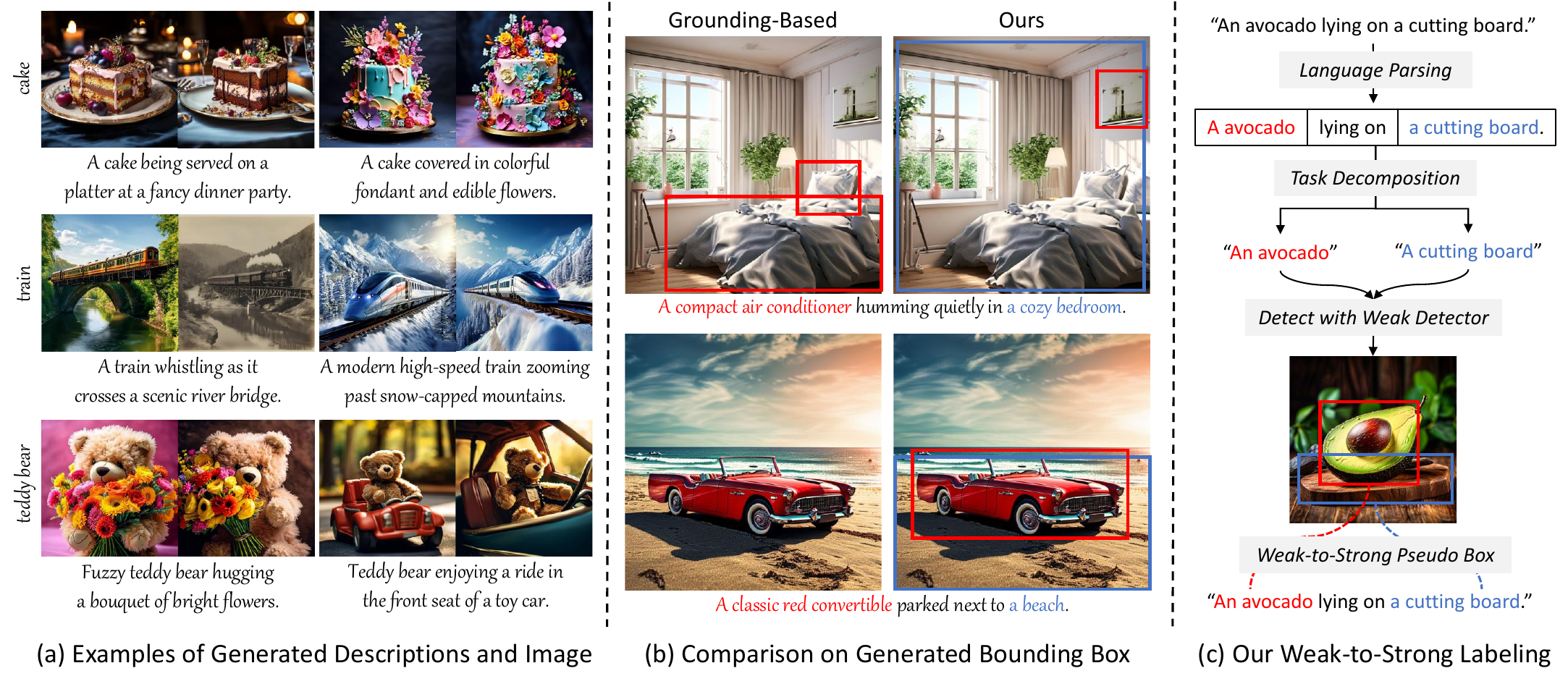}
    \captionsetup{font=footnotesize}
    \vspace{-6mm}
    \caption{
    (a) Qualitative examples of generated synthetic images and descriptions. (b) Comparison between grounding-based labeling and our weak-to-strong labeling. (c) Illustration of our weak-to-strong labeling, where we decompose the complex task into easy tasks. The bounding boxes collected from each task are combined to create strong compositional labels that train a strong detector. }
    \label{fig:qual}
    \vspace{-4mm}
\end{figure*}

\noindent\textbf{Generating Diverse Object Descriptions.}
We aim to generate a collection of dense image-text pairs, where a wide variety of visual entities is depicted in the image and text domain.
To achieve this, we initiate the process by generating diverse text descriptions for each entity with generative models. Recent advancements have demonstrated the remarkable capability of large language models (LLMs)~\cite{bert,GPT3,GPT4,touvron2023llama} to comprehend the real world in unprecedented detail. We capitalize on this knowledge by querying LLMs for plausible descriptions of objects under various conditions. For instance, we prompt an LLM with instructions such as, \textit{"Please list $\{ND\}$ plausible visual object descriptions for $\{class\}$ that are around $\{NW\}$ words in length. Consider incorporating diverse visual attributes, actions, and spatial or semantic relations with other objects in each description."} This approach allows us to efficiently gather prior knowledge about specific visual entities (\textit{i.e.} $\{class\}$), encompassing their likely attributes, natural co-occurrences with other objects, and the relationships between them. Representative examples are shown in Figs.~\ref{fig:framework} and \ref{fig:qual}.

The proposed LLM-based method for generating object descriptions is notable for its scalability and controllability. By adjusting parameters such as the pool size of visual entities (\ie, entity density), the number of descriptions (\textit{$\{ND\}$}) per entity (\ie, description density), and the length of each description (\ie, \textit{$\{NW\}$}), we can easily manage the diversity and volume of the generated descriptions. We borrow the pool of visual entities from well-curated lists of everyday object categories from popular object detection datasets~\cite{lin2014microsoft,objects365,gupta2019lvis}. The number of descriptions per entity is crucial for ensuring a comprehensive coverage of each entity's diversity, while the length of the descriptions influences the complexity of the resulting scenes. For example, longer descriptions tend to encompass more surrounding objects and intricate attributes, allowing us to tune the training samples' difficulty and quality.

\noindent\textbf{Generating Densely Paired Images with Diffusion Models.}
While previous work focuses on synthetic text augmentation~\cite{desco,dac}, our objective is to acquire densely paired image-text data in both image and text domains with text-to-image generative models. Diffusion-based text-to-image generation models~\cite{rombach2022high,ddpm,ddim} have recently demonstrated their capability to produce high-fidelity, photo-realistic images. The latest breakthroughs~\cite{sdxl,sdxl_turbo,pixart} in foundational diffusion models enable the generation of complex scenes featuring multiple objects with detailed descriptions. Our research investigates the extent to which these diffusion models can enhance the task of language-based object detection.

We condition the image generation process on generated object descriptions. It is different from previous methods~\cite{xie2023mosaicfusion} that used simple, hand-written prompts (e.g., “a photo of a [NAME]”). This approach allows us to explicitly introduce diversity by specifying the objects in the descriptions. As a byproduct, this strategy provides pairs of object descriptions and images for training purposes.

We investigate the impact of generating a diverse set of images from a single description (\ie, Image Density). By introducing varied initial noise into the diffusion model—achieved by manipulating random seeds—we generate different visuals of the same description. Examples of the variations are depicted in \figref{fig:qual}.

\noindent\textbf{Weak-to-Strong Pseudo Bounding Box Generation.}
Even if we have a collection of densely paired generated descriptions and images, accurate localization information of the depicted objects is crucial for training detectors on it. However, even recent pre-trained vision-language detectors often struggle to identify visual entities based on complex descriptions. Due to their compositional understanding capabilities, detectors like GLIP~\cite{glip} inaccurately localize or completely overlook objects, as illustrated in \figref{fig:qual}-(b, left). This issue presents a new challenge in utilizing generated data for training purposes.

To this end, we delve into strategies for achieving precise object localization using weak detectors (in terms of compositional understanding), thereby facilitating the generation of rich supervision for training stronger detectors. We term this as a weak-to-strong labeling method. Our key idea is simple and intuitive: we decompose the complex phrase grounding problem into multiple manageable detection tasks.
For this purpose, we make several key observations regarding the performance of recent visual-language detectors: 1) Although the detectors struggle to differentiate hard negative texts, they demonstrate proficiency in accurately localizing objects with positive texts (See the higher score for positive text (AP-dP) compared to overall detection Average Precision (AP-d) presented in \tabref{tab:sota}.) 2) The model performs better at detecting objects described with concise text rather than complex descriptions. (See a higher score for short descriptions (AP-dS) compared to long descriptions (AP-dL).

Guided by the observations, we reformulate the complex phrase grounding problem into multiple tractable detection tasks with positive and short descriptions. An overview of our weak-to-strong labeling approach is depicted in \figref{fig:qual}-(c). For each pair of generated images and object descriptions, we initiate the process by identifying all noun phrases with an NLP parser~\cite{spacy2}. We then treat each noun phrase as an independent description to detect the corresponding objects (task decomposition). This ensures satisfactory precision and recall, as demonstrated in \figref{fig:qual}-(b, right). Low-confidence predictions are filtered out based on a predetermined threshold p. The remaining predictions are re-assigned to the original position within the description, which results in a strong compositional label for the following step.

\begin{figure*}[t]
    \centering 
    \includegraphics[width=1\textwidth]{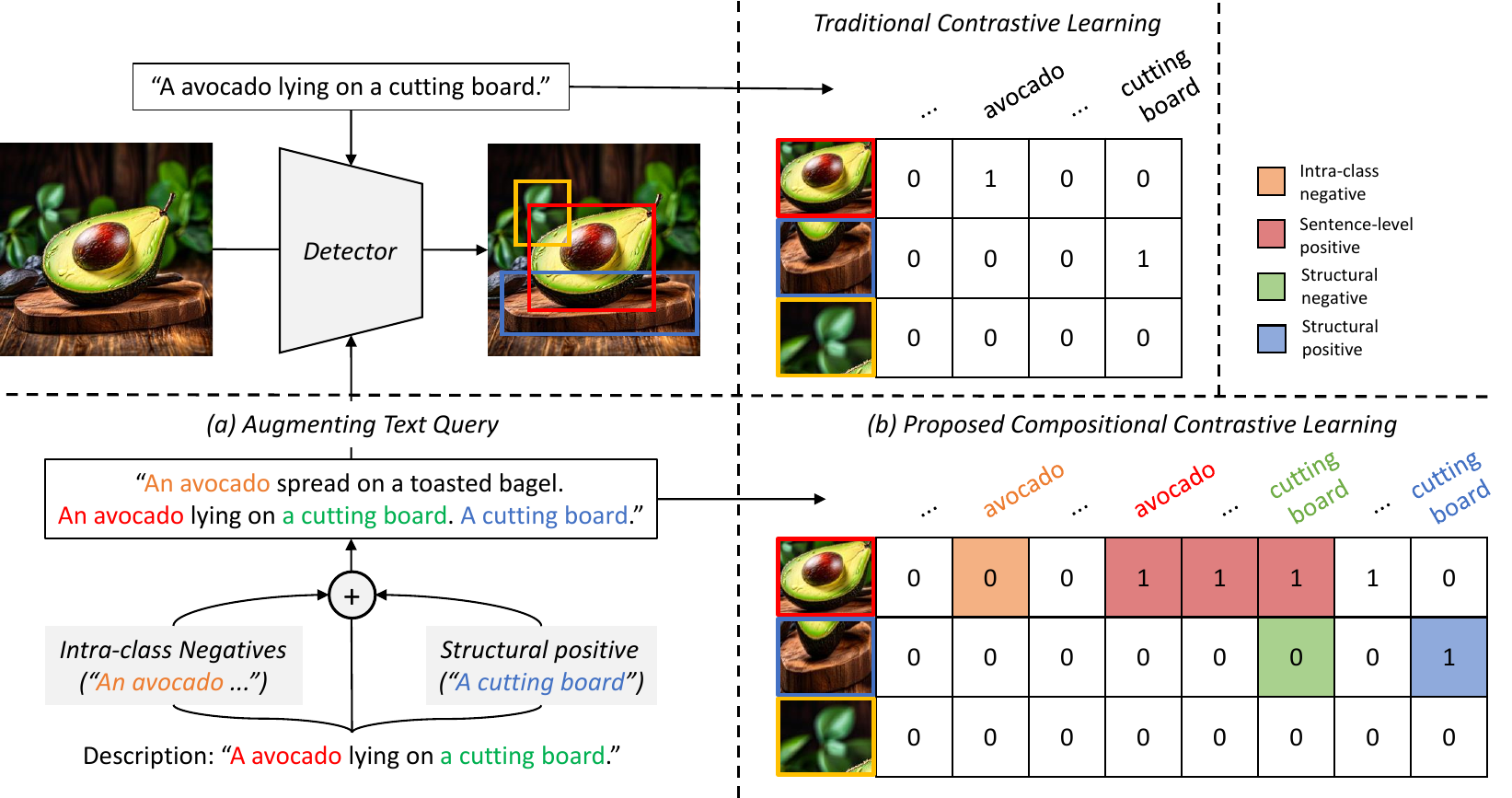}
    \captionsetup{font=footnotesize}
    \caption{
    Illustration of our compositional contrastive learning. (a) \textcolor{BurntOrange}{Intra-class negatives} from other images of the same class and \textcolor{RoyalBlue}{structural positives} are introduced to learn the context of descriptions. (b)  We associate the \textcolor{Maroon}{sentence-level positive} (\ie, the entire description sentence) with the pseudo bounding box of the ``\textcolor{red}{an avocado}'' while differentiating the \textcolor{Green}{structure negative} (\ie, the noun phrase ``\textcolor{Green}{a cutting board}'') from the pseudo bounding box of the  ``\textcolor{RoyalBlue}{a cutting board}''.}
    \vspace{-4mm}
    \label{fig:method}
\end{figure*}

\subsection{Description and Textural-structural Aware Compositional Contrastive Learning for Language-based Object Detection}
\label{method:training_w_synthetic}
A straightforward approach to utilize the generated triplets (image, object descriptions, bounding boxes) is to use as additional grounding data: learning the alignment between noun phrases and detected object regions. However, our preliminary investigations reveal that models naively trained with these triplets often exhibit degraded performance. This raises the question: How can we effectively learn from the generated samples? Analyzing representative failure cases (See \figref{fig:motivation}-(a)), we identify two critical functionalities for compositional understanding: description-awareness and textural-structural-awareness. We detail the methodologies for learning these functionalities using synthetic data and explore strategies to mitigate domain bias, thereby unlocking the synthetic data's full potential.

\noindent\textbf{Learning Description-awareness with Dense Contrastive Learning.} Traditional language-based detectors often lack description awareness, indiscriminately detecting entities, for example, detecting middle `women' regardless of the provided descriptions in \figref{fig:motivation}-(a). To address this, we introduce supervisory signals that lead the model to pay attention to the given descriptions. Specifically, we select intra-class negative captions from the description pool that belong to the same object category as the image and augment the input query Q with the negatives (\eg, ``\textcolor{BurntOrange}{An avocado} spread on a toasted bagel'' in \figref{fig:method}). Then the model is trained to disregard the visual entities for these negative captions. This approach demands that the model discerns between identical or similar noun phrases based solely on the context of entangled descriptions, significantly enhancing description-based detection accuracy. Notably, densely generated descriptions synergy well with this description-awareness training.

\noindent\textbf{Learning Textural-Structural-Awareness.} Existing language-based detectors often perform akin to a bags-of-words, indiscriminatively detecting all visual entities mentioned in the descriptions as detecting `black top' in the \figref{fig:motivation}-(a, left).
To overcome this, we aim to distinguish between subject and non-subject entities within descriptions. We use textural relation~\cite{spacy2} between noun phrases to identify subject and non-subject entities (i.e., visual entities within the descriptions). Then, the detector is instructed to ignore non-subject entities (\eg, ``lying on  \textcolor{Green}{a cutting board}'' in Fig.~\ref{fig:method})  based on the description. We term this concept as a structural negative. For the subject noun entity, we ensure that the entire positive descriptions are positively aligned (\textit{i.e.} sentence-level positive).
In addition, to prevent the model from taking shortcuts that overlook later nouns, we introduce structural positives (\eg, ``\textcolor{RoyalBlue}{A cutting board}'' in Fig.~\ref{fig:method}) by augmenting the model's textual input with the noun phrase of the non-subject entity. Then, the detector is trained to recognize the corresponding object for the structural positive query. Through this strategy, the model learns to differentiate identical noun phrases based on their structural role within the language query (subject vs. non-subject). 
This leads to significant improvements in performance, particularly for complex queries involving multiple visual entities.

\noindent\textbf{Preventing Domain Bias.}
While state-of-the-art diffusion models excel in producing high-quality, photo-realistic images, the synthesized images inevitably may exhibit artifacts. 
Moreover, even the most advanced diffusion models struggle to produce super-complex images with perfect text-to-image correspondence, leading to a loss of precise localization capabilities in complex scenes. This discrepancy raises the concern of language-based object detectors becoming overfitted to synthetic images, which could diminish their performance on real images.

To address these challenges, we propose two simple yet effective strategies: (1) Freezing the visual backbone while training detectors on synthesized data, which helps prevent the model's visual representations from overfitting to the synthetic distribution, and (2) Incorporating detection data as an additional training resource, which block the catastrophic forgetting of precise localization capabilities. These techniques collectively enable the model to seamlessly learn compositional visual language understanding without the risk of the domain gap.

\section{Experiments}

\subsection{Experimental Setting}
\noindent\textbf{Training Details.}
We base our proposals on two recent language-based object detectors: GLIP and FIBER. 
Specifically, we utilize the GLIP-T and FIBER-B versions, which employ Swin Transformer~\cite{liu2021swin} (Tiny and Base) as their visual backbones and BERT~\cite{bert} and RoBERTa~\cite{liu2019roberta} as their language backbones. 
We finetune their official weights using a combination of our generated datasets and the Objects365~\cite{objects365} object detection dataset. It should be noted that Objects365 has already been used in the training of both GLIP and FIBER. The inclusion of Objects365 aims to mitigate domain bias and preserve the detectors' innate ability to accurately localize objects within complex scenes, as detailed in the methods section. By default, in synthetic data generation, we use the category pool from Object365, ChatGPT3.5-Turbo~\cite{GPT3} for description generation, and Pixart~\cite{pixart} for image generation. For each category, we generate 20 descriptions and 8 images per description with different random seeds. In total, we generate 58,400 synthetic triplets. For additional details, please refer to the appendix.

\noindent\textbf{Evaluation Benchmarks.}
We benchmark our proposed approach on the OmniLabel~\cite{omnilabel} and D$^{3}$~\cite{d3} datasets, following their official evaluation protocols. These datasets provide a comprehensive evaluation of the language-based object detector's proficiency in detecting objects specified by complex descriptions. Unlike traditional benchmarks in referring expressions, these datasets introduce scenarios with descriptions that either refer to no object or to multiple instances in an image, thereby facilitating a detailed compositional understanding in language-based object detection tasks.

Both benchmarks offer a suite of sub-metrics designed for an in-depth analysis. Specifically, for OmniLabel, the Average Precision for categories (AP-categ) and for descriptions (AP-descr) quantify detection accuracy for standard plain object categories and for free-form textual descriptions, respectively. The overall metric, AP, is computed as the harmonic mean between AP-categ and AP-descr, providing a balanced measure of both performances. Further dissecting description-based performance, the AP-descr-pos metric isolates the evaluation to positive descriptions, while AP-descr-S/M/L categorizes performance metrics according to the length of the descriptions (short, medium, and long), offering detailed insights into the detection efficacy relative to description complexity.
The D$^{3}$ dataset categorizes descriptions into ABS (``absence'') and PRES (``presence'') based on whether the description includes expressions of absence (e.g., ``without''). In addition to an overall evaluation metric encompassing all descriptions (referred to as FULL), D$^{3}$ provides distinct metrics for ABS and PRES.

\begin{table*}[t!]
    \centering
\setlength{\tabcolsep}{5pt}
{\renewcommand\arraystretch{0.95}
    \resizebox{\textwidth}{!}
    {
        \begin{tabular}{lc|ccccccc|ccc}
        \toprule
        \multicolumn{1}{c}{} &\multicolumn{1}{c|}{} & \multicolumn{7}{c|}{\textbf{OmniLabel~\cite{omnilabel}}} & \multicolumn{3}{c}{\textbf{D$^3$~\cite{d3}}}\\ 
        \multicolumn{1}{l}{Model} & Backbone & AP & AP-c & AP-d & AP-dP & AP-dS & AP-dM & AP-dL & Full & Pres & Abs\\
        \toprule
            RegionCLIP~\cite{zhong2022regionclip}   & ResNet-50   &2.7 & 2.7   & 2.6   & 3.2 & 3.6 & 2.7 & 2.3 & - & - &- \\
            Detic~\cite{zhou2022detecting}        & Swin-B      &8.0 & 15.6  & 5.4   & 8.0 & 5.7 & 5.4 & 6.2 & - & - & -\\
            Grounding-DINO~\cite{groundingdino}   & Swin-B      & - & - & - & - & - & - & - & 20.7 & 20.1 & 22.5 \\
            OFA-DOD~\cite{d3}          & Swin-B      & - & - & - & - & - & - & - & 21.6 & 23.7 & 15.4 \\
            \midrule 
            \midrule 
            GLIP-T~\cite{glip}       & Swin-T      &19.3 & 23.6 & 16.4  & 25.8 & 29.4 & 14.8 & 8.2 & 19.1 & 18.3 & 21.5\\
            \CCG w/ Ours              &\CCG Swin-T      &\CCG  \textbf{24.3} & \CCG \textbf{23.9} &\CCG  \textbf{24.7}  &\CCG  \textbf{34.4} &\CCG  \textbf{39.3} &\CCG  \textbf{21.6} &\CCG  \textbf{16.4} &\CCG  \textbf{26.0} &\CCG  \textbf{25.6} &\CCG  \textbf{27.1}\\ 
            \midrule 
            FIBER-B~\cite{fiber}      & Swin-B      &25.7 & 30.3 & 22.3  & 34.8 & 38.6 & 19.5 & 12.4 & 22.7 & 21.5 & 26.0\\ 
            \CCG w/ Ours              & \CCG Swin-B      &\CCG \textbf{30.5} & \CCG \textbf{31.6}	&\CCG  \textbf{29.5}	& \CCG \textbf{40.3} &\CCG  \textbf{43.7} &\CCG  \textbf{26.3} &\CCG  \textbf{21.3}  &\CCG  \textbf{26.5} &\CCG  \textbf{26.0} &\CCG  \textbf{27.7}\\ 
            \midrule 
            \midrule 
            Desco-GLIP~\cite{desco}   & Swin-T      &23.8 & \textbf{27.4}  & 21.0  & 30.3 & 33.7 & 19.0 & 13.7 & 24.2 & 22.9 & 27.8\\ 
            \CCG w/ Ours              &\CCG  Swin-T      &\CCG \textbf{26.5} &\CCG  27.1  &\CCG  \textbf{25.9}  &\CCG  \textbf{35.6} &\CCG  \textbf{38.1} &\CCG  \textbf{23.2} &\CCG  \textbf{18.7} &\CCG  \textbf{29.3}	&\CCG  \textbf{29.1}	&\CCG  \textbf{30.1}\\ 
            \midrule 
            Desco-FIBER~\cite{desco}  & Swin-B      &29.3 & 31.6 & 27.3  & 37.7 & 42.8 & 24.4 & 18.6 & 28.1 & 27.2 & \textbf{30.5}\\ 
            \CCG w/ Ours              &\CCG  Swin-B      &\CCG \textbf{32.0} &\CCG \textbf{33.1}	&\CCG \textbf{30.9} &\CCG \textbf{40.4} &\CCG \textbf{45.2} &\CCG \textbf{27.7} &\CCG \textbf{22.9} &\CCG \textbf{30.8} &\CCG  \textbf{31.0} & \CCG 30.4\\
        \bottomrule
        \end{tabular}
    }
}
\captionsetup{font=footnotesize}
\caption{Performance comparison with state-of-the-art methods. We apply our method on top of diverse existing methods and significantly boost the performance.}
\label{tab:sota}
\vspace{-6mm}
\end{table*}

\subsection{Main Results}
We evaluate the impact of the proposed learning framework with the densely generated triplets. Experimental results on OmniLabel and D$^{3}$ benchmarks are summarized in \tabref{tab:sota}. We first finetune two baseline models, GLIP and FIBER, and observe significant enhancements in language-based object detection performance across both datasets. This implies that the proposed learning framework is generic over different detection architectures and evaluation scenarios.  Notably, the GLIP model's performance shows a substantial improvement, with an increase of +5.0AP and +6.9AP on the overall metrics for the OmniLabel and D$^{3}$ datasets, respectively. The enhancements are particularly pronounced for long queries (\textit{i.e.}, AP-dL in OmniLabel), where the performance of the GLIP model doubles from 8.2 to 16.4.

We then explore the synergy between our proposals and the prior language augmentation-based method (\ie, DesCo~\cite{desco}). In this configuration, we apply their methods to enrich the language queries within the detection dataset during training. As shown in the table, our proposal surpasses their models, DesCo-GLIP and DesCo-FIBER, by a considerable margin across both datasets. This shows that augmenting solely within the textual domain is insufficient. Our compositional contrastive learning on densely generated triplets offers distinct and substantial improvements.

\subsection{Ablation Study and Analysis}

To assess the impact of our proposed components, we conduct comprehensive ablation studies on the FIBER-B model.

\begin{wraptable}{r}[.001in]{0.48\textwidth}
\vspace{-8ex}
\resizebox{0.48\textwidth}{!}{
    
     \setlength{\tabcolsep}{3pt}
     \def\arraystretch{1.6}
     \begin{tabular}{l|  c c c | c c c c}
        \toprule
        learning method      & AP & AP-c     & AP-d   & AP-dp & AP-dS & AP-dM & AP-dL \\
        \midrule
        FIBER-B                & 25.7 & 30.3 & 22.3  & 34.8 & 38.6 & 19.5 & 12.4 \\ 
        \midrule
        Gen-only               & 25.5 & 27.7 & 23.7	& 34.4	& 41.5	& 19.6	& 12.4 \\
        \midrule
        (+) Det data           & 26.3 & 30.2 & 23.3	& 34.2	& 41.0	& 19.7	& 11.5 \\
        (+) Freeze vis-back    & 26.8 & 31.3 & 23.4	& 34.4	& 40.8	& 19.5	& 11.8 \\
        \midrule
        (+) Intra-neg          & 29.0 & 30.9 & 27.4	& 36.6	& 44.2	& 24.0	& 14.9 \\
        \midrule
        (+) Struct-neg          & 29.0 & 31.0	& 27.3	& 37.1	& 43.7	& 24.4	& 16.2 \\
        (+) Struct-pos          & 30.5 & 31.6	& 29.5	& 40.3	& 43.7	& 26.3	& 21.3 \\
        \bottomrule
        \end{tabular}
     } 
     \vspace{-2ex}
     \caption{Ablation on compositional contrastive learning.}
     \vspace{-6.ex}
     \label{tab:learning_method}
\end{wraptable}

\noindent\textbf{Effective learning signals with synthetic data.} We validate the impact of the proposed learning methods. Experimental results are summarized in \tabref{tab:learning_method}. We start by naive finetuning only on the densely generated triplets: treating these triplets similarly to conventional grounding data. (\ie, Gen-only). While the description-based performance is improved, the precise localization capability with the given plain category is largely degraded. To mitigate the detrimental effects of the distributional discrepancies between generated and real-world data, we employ common object detection datasets~\cite{objects365} as a form of regularization and freeze the visual backbone during training. As shown in the table, each learning technique helps to maintain or even improve precise localization capability and thus enables solid learning from the synthetically generated datasets. Next, we explore the impact of the proposed contrastive learning methods. By contrasting dense descriptions from the same visual entity (\ie, Intra-neg), the model faithfully learns the description awareness, leading to the significant improvements of 4.0AP in the description-based performance. We then explore the text structural-based contrastive learning. Naively treating the non-subject object as negative for the description doesn’t bring notable improvements (\ie, Struct-neg). However, when the concept of structural positive is included, the model is enforced to discriminate the same phrases according to their structural role in the description. This greatly improves description-based performance, especially the notable gain of 6.4AP for long queries. To sum up, all the proposed learning methods show their unique effect and the performance improvements of the final model over the baseline are significant.

\begin{table*}[t!]
\centering
\subfloat[\scriptsize Entity Diversity]{
     \label{tab:den_ent}
     \resizebox{0.31\textwidth}{!}{
     \setlength{\tabcolsep}{3pt}
     \def\arraystretch{1.6}
     \begin{tabular}{l|  c c c | c c}
        \toprule
        category               & AP & AP-c  & AP-d  & AP-dS & AP-dL \\
        \midrule
        COCO (80)              & 29.7	&31.0 & 28.5 &43.8 & 18.6  \\ 
        \textbf{O365 (365)}    & 30.5	&31.6 &	29.5 &43.7 & 21.3 \\ 
        LVIS (1203)            & 31.1	&30.9 & 31.3 &45.3 & 23.3  \\ 
        \bottomrule
        \end{tabular}
     }
 }
\subfloat[\scriptsize Description Density]{
     \label{tab:den_des}
     \resizebox{0.31\textwidth}{!}{
     \setlength{\tabcolsep}{3pt}
     \def\arraystretch{1.6}
     \begin{tabular}{l|  c c c | c c}
        \toprule
        num. des.               & AP & AP-c     & AP-d  & AP-dS & AP-dL \\
        \midrule
        5 per ent.             & 29.1 & 31.0 & 27.5	& 42.2 & 17.4   \\ 
        10 per ent.            & 29.7 & 31.1 & 28.4	& 43.6 & 18.4   \\ 
        \textbf{20 per ent.}   & 30.5 &31.6  & 29.5 & 43.7 & 21.3 \\  
        \bottomrule
        \end{tabular}
     }
}
\subfloat[\scriptsize Image Density]{
     \label{tab:den_seed}
     \resizebox{0.31\textwidth}{!}{
     \setlength{\tabcolsep}{3pt}
     \def\arraystretch{1.6}
     \begin{tabular}{l|  c c c | c c}
        \toprule
        num. img.               & AP & AP-c     & AP-d   & AP-dS & AP-dL \\
        \midrule
        2 per des.            & 29.8 & 31.2 & 28.6 & 43.1 & 19.3\\ 
        4 per des.            & 29.7 & 31.3 & 28.2 & 42.0 & 19.2 \\ 
        \textbf{8 per des.}   & 30.5 &31.6  & 29.5 & 43.7 & 21.3   \\  
        \bottomrule
        \end{tabular}
     }
}

\vspace{-2mm}
\captionsetup{font=footnotesize}
\caption{
Analysis on scaling factors for generated triplets.} 
\label{tab:abl_des_gen}
\vspace{-8mm}
\end{table*}

\noindent\textbf{Scaling factors for the generated dataset.} The scale of a dataset is a crucial determinant of its effectiveness. We investigate various design choices that influence the size of the generated datasets, identifying the critical factors for efficient data scaling. We mainly explore three factors: density of entity, description, and image.

We first study the density of the covered entity by scaling the category set. We borrow a well-curated list of classes from COCO~\cite{coco}, Object365~\cite{objects365} and LVIS~\cite{lvis}.
We generate dense synthetic triplets for each set and use them to train a detector. As shown in \tabref{tab:den_ent}, the description-based performance gradually improved as the scale of the visual entity grew. This implies that it is crucial to learn from dense triplets of diverse visual entities. On the contrary, for the plain-category name-based detection, the set of the Object365 class shows the best performance. This is because existing diffusion models also suffer from long-tailed issues and have trouble illustrating uncommon objects. Considering the balance between AP-c and AP-d, we use the category pool of Object365 as the default for other experiments. Our default setting is noted in bold. 

We also explore the number of generated descriptions for each visual entity. We vary the number from 5 to 20 and report the performance of the detector trained on corresponding generated triplets in \tabref{tab:den_des}. The number of descriptions per entity greatly impacts overall scores, especially on the long query. This shows the importance of dense triplets and highlights the potential of an easy-to-scalable synthetic data generation framework.

Lastly, we study whether the density of generated images matters for the efficiency of the framework. To generate diverse images for a given description, we generate multiple variations by introducing different initial noises into the diffusion models, achieved by varying the random seed. We adjust the number of random seeds used for image generation from 2 to 8. As indicated in \tabref{tab:den_seed}, the diversity of images proves beneficial. The model benefits from learning across multiple visual variations of a single description, leading to a robust alignment between visual and linguistic representations.

\begin{wraptable}{r}[.001in]{0.48\textwidth}
\vspace{-6ex}
\resizebox{0.48\textwidth}{!}{
    
     \setlength{\tabcolsep}{3pt}
     \def\arraystretch{1.6}
     \begin{tabular}{l|  c c c | c c c c}
        \toprule
        strategy                                   & AP & AP-c     & AP-d   & AP-dp & AP-dS & AP-dM & AP-dL \\
        \midrule
        Grounding-based             &29.3 & 31.3 & 27.5	& 37.4	& 43.4	& 24.1	& 16.2 \\ 
        \textbf{Weak-to-Strong}  & 30.5	&31.6 &	29.5 & 40.3 &43.7 & 26.3 & 21.3 \\ 
        \bottomrule
        \end{tabular}
     }
     \vspace{-2ex}
     \caption{Ablation on pseudo label generation strategies.}
     \label{tab:bbox_gen}
     \vspace{-5ex}
\end{wraptable}

\noindent\textbf{Pseudo box generation strategy.}
We study the impact of the pseudo bounding box generation strategy on the final performance. As shown in Table~\ref{tab:bbox_gen}, the proposed weak-to-strong method brings notable improvements compared to conventional grounding-based technique. This shows the importance of the quality of the bounding box for compositional learning.

\begin{wraptable}{r}[.001in]{0.48\textwidth}
    \vspace{-5ex}
    \resizebox{0.48\textwidth}{!}{
     \setlength{\tabcolsep}{3pt}
     \def\arraystretch{1.6} 
     \begin{tabular}{l|  c c c | c c | c c}
        \toprule
         des. length        & AP & AP-c & AP-d   & AP-dS & AP-dL & NOUN & ADJ\\
        \midrule
        6 words             &29.7  & 31.0 &28.5	 & 43.8 &18.6 & 2.89 & 0.87 \\ 
        8 words             &30.3  & 31.2 & 29.3 &43.0 & 20.5 & 3.29 & 1.04 \\ 
        \textbf{10 words}   &30.5	&31.6 &	29.5 &43.7 & 21.3 & 4.10 & 1.49 \\  
        12 words            &29.9	&31.5 & 28.5 &43.8 & 18.3 & 4.24 & 1.52 \\ 
        \bottomrule
        \end{tabular}
     }
     \vspace{-2ex}
     \caption{Additional analysis on the effective length of descriptions.}
     \label{tab:des_length}
     \vspace{-5ex}
\end{wraptable}

\noindent\textbf{Effective description length.} As highlighted in the methods section, the specified length of the descriptions affects the complexity of the object descriptions and the resultant images. To demonstrate this concept, we adjust the requested description lengths from 6 to 12 words and conduct a textural analysis with an NLP parser~\cite{loper2002nltk}.
In Table~\ref{tab:des_length}, we report the average number of nouns and adjectives per description, which correlates with the number of objects and their specified attributes, respectively. 
Monotonic increased factors over the description length show a positive correlation between the requested description length and scene complexity.

We then evaluate the effectiveness of our learning framework as the complexity of the generated image/text combinations varies. 
Although our approach performs robustly across all description lengths, optimal results were observed at a description length of 10 words.
Short descriptions tend to generate overly simplistic descriptions and images, which are insufficient for learning nuanced description and structure sensitivity. Conversely, longer descriptions risk exceeding the capabilities of state-of-the-art models, potentially leading to generating images that are more likely to contain artifacts, such as missing objects or inaccurately depicted attributes. This may bring noise in the language-based object detector training.

\noindent\textbf{Additional analyses.} We present further ablation studies and analyses on various factors, such as pseudo box generation strategy, frozen backbones, the choice of diffusion models, and the efficiency of our framework. Additionally, we include qualitative detection results from our models and others in the supplementary materials.

\section{Conclusion}

Although vision-language (VL) models have made notable progress in language-based object detection, they continue to face challenges in comprehensively understanding the compositions of visual scenes and textual descriptions. This leads to a noticeable decline in performance when faced with complex language queries. To our knowledge, we first propose to automatically generate synthetic triplets containing diverse and complex text descriptions, corresponding images, and reliable pseudo-bounding boxes. These synthetic triplets lead a VL model to learn compositional capability with our proposed compositional contrastive learning. Our approach is model-agnostic, which can be applied to improve diverse existing VL models and significantly boost the performance on this challenging task.

\section*{Acknowledgement}
This study was supported by the following grants: the Institute of Information \& communications Technology Planning \& Evaluation(IITP) grant funded by the Korea government(MSIT) (No. 2022-0-00871, Development of AI Autonomy and Knowledge Enhancement for AI Agent Collaboration, 70\%), (No. RS-2019-II190079, Artificial Intelligence Graduate School Program(Korea University), 5\%), the National Research Foundation of Korea(NRF) grant funded by the Korea government(MSIT)(No. RS-2024-00341514, 15\%), Culture, Sports and Tourism R\&D Program through the Korea Creative Content Agency grant funded by the Ministry of Culture, Sports and Tourism in 2024(Project Name: International Collaborative Research and Global Talent Development for the Development of Copyright Management and Protection Technologies for Generative AI, Project Number: (RS-2024-00345025, 10\%)

\appendix
\addcontentsline{toc}{section}{Appendices}

\section*{Appendices}

This supplementary material contains more details including:
\begin{enumerate}[label=\Alph*.]
    \item Additional ablation study and analysis,
    \item Limitations of our work,
    \item Qualitative comparisons.
\end{enumerate}

\section{Additional ablation study and analysis}

\noindent\textbf{Pseudo box generation strategy.}
As shown in the main paper, the strategy for generating pseudo-bounding boxes significantly influences the overall performance, with our proposed weak-to-strong methods yielding remarkable enhancements. For a more detailed understanding, we provide more experimental comparisons and analyses in this section.

\begin{table*}[b!]
 \centering
 \subfloat[\scriptsize pseudo box generation strategy]{
     \label{tab:supple_bbox_gen}
     \resizebox{0.48\textwidth}{!}{
     \setlength{\tabcolsep}{3pt}
     \def\arraystretch{1.6}
     \begin{tabular}{l|  c c c | c c | c}
        \toprule
        strategy                    & AP & AP-c     & AP-d   & AP-dS & AP-dL & Qual \\
        \midrule
        Grounding-based             &29.3 & 31.3 & 27.5	& 43.4	& 16.2 & 53.8 \\ 
        \textbf{Weak-to-Strong}  & 30.5	&31.6 &	29.5 &43.7  & 21.3 & 65.0 \\ 
        
        \bottomrule
        \end{tabular}
     }
 }
 \subfloat[\scriptsize confidence threshold]{
     \label{tab:supple_bbox_conf}
     \resizebox{0.48\textwidth}{!}{
     \setlength{\tabcolsep}{3pt}
     \def\arraystretch{1.6}
     \begin{tabular}{c|  c c c | c c | c}
        \toprule
        confidence threshold       & AP & AP-c     & AP-d  & AP-dS & AP-dL & Recall \\
        \midrule
        0.3             & 29.7	&31.5 & 28.2 &41.9 & 18.9 & 0.99 \\ 
        \textbf{0.5}    & 30.5	&31.6 &	29.5 &43.7 & 21.3 & 0.90 \\ 
        0.7             & 29.6	&30.9 & 28.4 &42.8 & 19.2 & 0.53 \\ 
        
        \bottomrule
        \end{tabular}
     }
 }
\captionsetup{font=footnotesize}
\caption{
Additional ablation on pseudo label generation strategies.}
\label{tab:supple_abl_box}
\end{table*}

We first assess the quality of various pseudo bounding boxes.
Evaluating the quality of pseudo bounding boxes on a large set of synthetic images is challenging due to the absence of ground truth detection labels. 
For this reason, we manually annotated 100 randomly selected synthetic images and conducted direct evaluations of various pseudo bounding boxes. 
Our weak-to-strong method significantly improves the quality of bbox upon the grounding-based baseline, from 53.8AP to 65.0AP, with absolutely high accuracy (\ie, See Qual in \tabref{tab:supple_bbox_gen}).

We further examine the impact of the thresholding hyperparameter p, which is used to filter out predictions with low confidence, as described in the main paper. We adjust p within the range of 0.3 to 0.7. As shown in \tabref{tab:supple_bbox_conf}, optimal performance is observed at a threshold of 0.5, achieving a high recall rate for visual entities. Here, we treat noun phrases in descriptions as distinct visual entities and quantify their recall rate in the pseudo boxes. A higher parameter results in the exclusion of most predictions, leading to a significantly reduced recall rate. Conversely, setting the lower threshold increases the recall rate but also introduces noisy predictions into the pseudo labels, hindering the effectiveness of the learning process.

\begin{table*}[t!]
 \centering
\subfloat[\scriptsize Diffusion Model]{
     \label{tab:supple_diff_model}
     \resizebox{0.48\textwidth}{!}{
     \setlength{\tabcolsep}{3pt}
     \def\arraystretch{1.6}
     \begin{tabular}{l|  c c c | c c c c}
        \toprule
        diff. model                                   & AP & AP-c     & AP-d   & AP-dp & AP-dS & AP-dM & AP-dL \\
        \midrule
        \textbf{Pixart}     &30.5 &31.6  & 29.5 & 40.3 &43.7 & 26.3 & 21.3 \\
        SDXL                &30.3 &31.2	 & 29.4	& 39.8 & 44.7 & 26.3 & 20.0  \\ 
        SDXL-Turbo          &29.9 &31.0	 & 28.9	& 39.5 & 43.5 & 25.7 & 19.9  \\ 
        \bottomrule
        \end{tabular}
     }
 }
\subfloat[\scriptsize Language Model]{
     \label{tab:supple_lang_model}
     \resizebox{0.48\textwidth}{!}{
     \setlength{\tabcolsep}{3pt}
     \def\arraystretch{1.6}
     \begin{tabular}{l|  c c c | c c c c}
        \toprule
        lang. model                                   & AP & AP-c     & AP-d   & AP-dp & AP-dS & AP-dM & AP-dL \\
        \midrule
        llama-70b               & 30.2 &31.0  & 29.3 & 40.3	&44.0 & 26.2 & 19.8  \\ 
        \textbf{GPT3.5-turbo}   & 30.5 &31.6  & 29.5 & 40.3 &43.7 & 26.3 & 21.3 \\ 
        GPT4                    & 30.6 &31.6  & 29.7 & 40.7	&44.2 & 26.5 & 20.8  \\ 
        \bottomrule
        \end{tabular}
     }
 }
\captionsetup{font=footnotesize}
\vspace{-2.ex}
\caption{
Additional analysis on choice of (a) the diffusion model and (b) the language model.}
\label{tab:abl_img_gen}
\vspace{-4.ex}
\end{table*}

\noindent\textbf{Choice of the diffusion model.}
We explore how the choice of text-to-image model influences the final performance of object detection. In this evaluation, we explore three state-of-the-art text-to-image models: Pixart~\cite{pixart}, SDXL~\cite{sdxl}, and SDXL-Turbo~\cite{sdxl_turbo}. Using these models, we generate varied sets of images for identical object descriptions, resulting in different collections of densely paired synthetic triplets. These triplets are then utilized to train the FIBER-B model and the experimental results are summarized in \tabref{tab:supple_diff_model}. Our learning framework reliably enhances performance across the model, though the diffusion models exhibit variable results in terms of the visual quality of generated images and the accuracy of image-text correspondence. This highlights the robustness of our approach regardless of the diffusion model chosen. Pixart is selected as our default setting due to its marginally superior performance and fast inference speed.

\noindent\textbf{Choice of language model.}
We investigate the impact of selecting different large language models (LLMs) on object detection performance. In this study, we evaluate three LLMs: LLaMA2-70B~\cite{touvron2023llama}, ChatGPT-3.5 Turbo~\cite{GPT3}, and ChatGPT-4~\cite{GPT4}. Similar to the above experiments, we generate varied collections of densely paired synthetic triplets and use them to train the detectors. The results are summarized in the table. Although superior language models slightly show improvements, the performance differences among them are marginal. Taking into account both performance and inference efficiency, we choose ChatGPT-3.5 Turbo as a default setting.

\noindent\textbf{Freezing network components.}
In our main paper, we propose that freezing the visual backbone helps to prevent the model from overfitting to the synthetic distribution during training. To substantiate this claim more convincingly, we conduct a thorough exploration into the effects of freezing different components of the detector. Common language-based object detectors are comprised of three key components: 1) a visual backbone for understanding the input image, 2) a language backbone for extracting linguistic features, and 3) fusion layers that fuse information from both modalities to detect objects according to the text query. We experiment with freezing each component individually and assess the impact on performance compared to a baseline model that is naively trained on generated triplets and the Objects365~\cite{objects365} detection dataset.

\begin{wraptable}{r}[.001in]{0.48\textwidth}
\vspace{-4ex}
\resizebox{0.48\textwidth}{!}{
    
     \setlength{\tabcolsep}{3pt}
     \def\arraystretch{1.6}
     \begin{tabular}{l|  c c c | c c c c}
        \toprule
        learning method      & AP & AP-c     & AP-d   & AP-dp & AP-dS & AP-dM & AP-dL \\
        \midrule
        w/o freeze  & 26.3 & 30.2 & 23.3	& 34.2	& 41.0	& 19.7	& 11.5 \\
        \midrule
        \textbf{Freeze Vis.}  & 26.8 & 31.3 & 23.4	& 34.4	& 40.8	& 19.5	& 11.8 \\
        Freeze Lang.          & 26.1	& 30.6	& 22.8	& 35.9	& 38.5	& 19.6	& 12.1 \\
        Freeze Fuse.    & 26.4	& 30.1	& 23.5	& 34.5	& 41.2	& 19.9	& 11.7 \\
        \bottomrule
        \end{tabular}
     } 
     \vspace{-2ex}
     \caption{Additional ablation on freezing network components.}
     \vspace{-4.ex}
     \label{tab:supple_freeze}
\end{wraptable}

The results, as presented in Table~\ref{tab:supple_freeze}, indicate that freezing the visual backbone yields better performance than freezing the other components or not applying any freezing technique at all (\ie, w/o freeze). Moreover, freezing the language backbone shows degraded performance, particularly in description-based object detection. This reveals that the pre-trained image representations may generalize well, whereas the bottlenecks lie in the language component. Furthermore, compositional learning with synthetic triplets may degrade the generality of visual representation. Therefore, the optimal strategy is to teach the model to understand complex language queries while reading out high-quality pre-trained visual representations (\ie, freezing visual backbone) for better compositional understanding.

\noindent\textbf{Efficiency of the framework.}
Our framework brings minimal training costs. 
The generation of descriptions, images, and bounding boxes takes a total of 7.5 hours (0.5 hr, 6 hr, 1 hr for each) for 58K triplets, and the additional training requires only 3 hours. 
These costs are efficient, especially compared to the significant data curation cost of 1.3M data and the 72 hours of training time required for GLIP~\cite{glip}. 
Our efficient framework supports the extension of data generation processes for novel classes. 
Most importantly, our framework significantly enhances detector performance for both novel classes (not covered in the data generation) and complex object descriptions, even with a relatively small number of generated triplets.

\section{Limitations of our work}
While our framework significantly enhances the compositional understanding of language-based object detectors, there are several limitations within our proposal that could be interesting points for future research.

Firstly, despite our efforts to mitigate the effects of artifacts in generated triplets—such as freezing the visual backbone, employing real detection data as a regularizer, and training exclusively with detectable objects—implementing more sophisticated filtering techniques to exclude low-quality samples could be beneficial. The criteria for ``low-quality'' can vary, encompassing aspects like visual quality~\cite{kirstain2024pick,wu2023human} and the accuracy of image-text correspondence~\cite{hessel2021clipscore,huang2024t2i}. Exploring the potential synergy between various filtering methods and our framework could yield insights, similar to previous studies~\cite{gadre2024datacomp}.

\begin{figure*}[t]
    \centering 
    \includegraphics[width=1\textwidth]{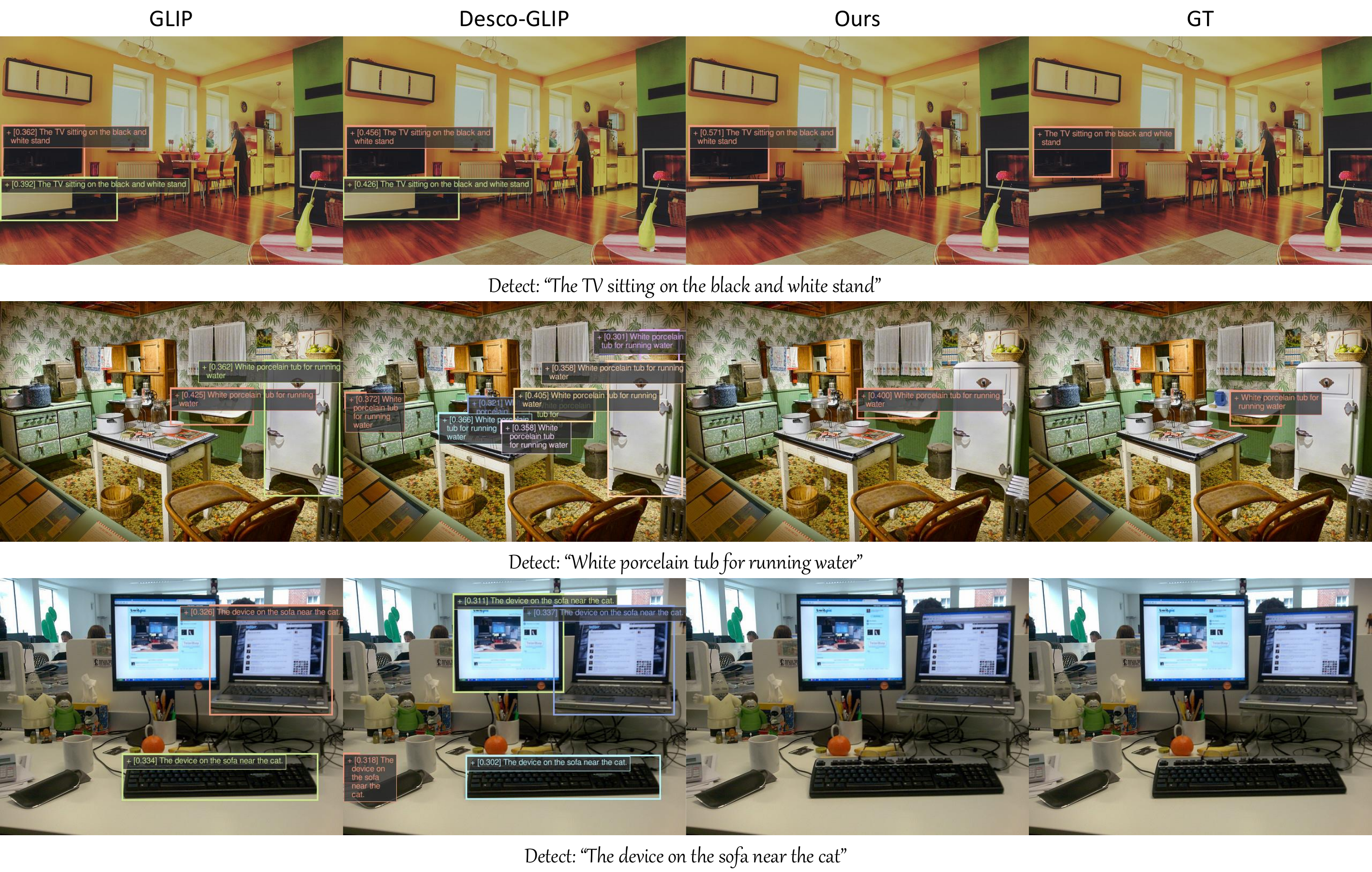}
    \captionsetup{font=footnotesize}
    \caption{
    Qualitative comparisons on OmniLabel~\cite{omnilabel} benchmark. We can observe clear improvements in compositional understanding against GLIP~\cite{glip} and Desco-GLIP~\cite{desco}.}
    \label{fig:qual_glip}
\end{figure*}

\begin{figure*}[t]
    \centering 
    \includegraphics[width=1\textwidth]{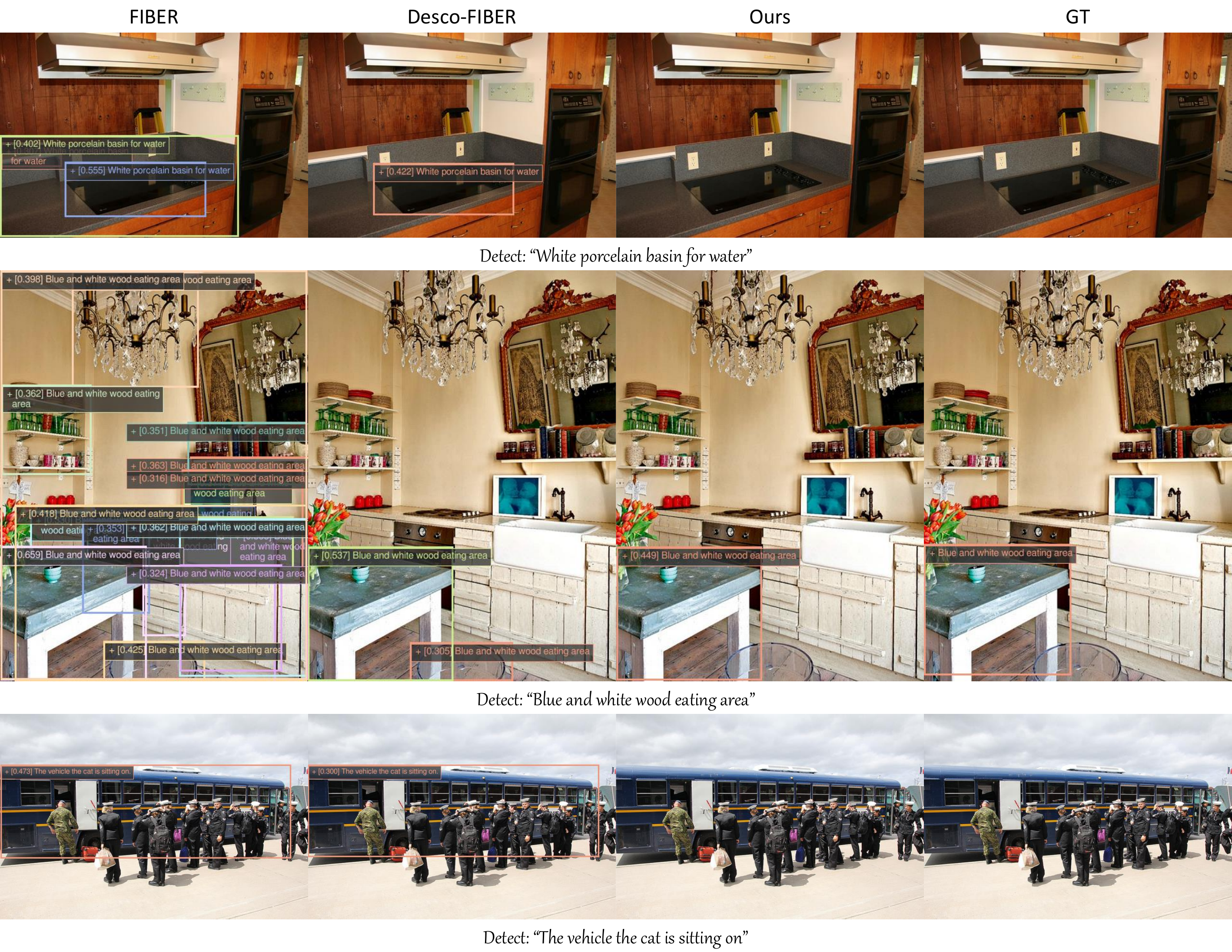}
    \captionsetup{font=footnotesize}
    \caption{
    Qualitative comparisons on OmniLabel~\cite{omnilabel} benchmark. We can observe clear improvements in compositional understanding against FIBER~\cite{fiber} and Desco-FIBER~\cite{desco}.}
    \label{fig:qual_fiber}
\end{figure*}

Moreover, while we instruct Large Language Models (LLMs)~\cite{GPT3,GPT4} to generate plausible descriptions of visual entities under a broad range of conditions, these prompts may not encompass all types of textual expressions. For example, LLMs typically describe objects based on their features but might not employ negations~\cite{hosseini2021understanding} (e.g., "A dog \textit{without} dots"). Although our model demonstrates strong generalization capabilities regarding the concept of negation (See improved Abs scores in Table 1 of the main paper), curating synthetic triplets aimed at such specific cases could further enhance performance.

\section{Qualitative comparisons}
In this section, we present qualitative comparisons against previous methods. The \figref{fig:qual_glip} compares our model with the pre-trained GLIP~\cite{glip} and the language-augmentation-based method, Desco-GLIP~\cite{desco}. Additionally, \figref{fig:qual_fiber} provides qualitative comparisons between our model, FIBER~\cite{fiber}, and Desco-FIBER~\cite{desco}.
As illustrated in both figures, our model successfully identifies and locates the target object only, disregarding any confusable objects in the image based on the given descriptions.

\bibliographystyle{splncs04}
\bibliography{main}
\end{document}